\documentclass[lettersize,journal]{IEEEtran}
\usepackage{amsmath,amsfonts}
\usepackage{algorithmic}
\usepackage{algorithm}
\usepackage{array}
\usepackage{subfigure}
\usepackage{textcomp}
\usepackage{stfloats}
\usepackage{booktabs}
\usepackage{multirow}
\usepackage{hyperref}
\usepackage{multicol}
\usepackage{bigstrut}
\usepackage{url}
\usepackage{makecell}
\usepackage{amssymb}
\usepackage{verbatim}
\usepackage{graphicx}
\usepackage{cite}
\usepackage{pifont}
\usepackage{color}
\usepackage[table]{xcolor}
\usepackage{colortbl}
\usepackage{soul}
 
\begin{document}

\title{Embedding Generalized Semantic Knowledge into Few-Shot Remote Sensing Segmentation}

\author{Yuyu Jia, 
Wei Huang, 
Junyu Gao, ~\IEEEmembership{Member,~IEEE,}
\\
Qi Wang, ~\IEEEmembership{Senior Member,~IEEE}
Qiang Li,  ~\IEEEmembership{Member,~IEEE,}

\thanks{This work was supported by the Natural Science Foundation of Shaanxi Province under Grant 2021JC-15 and the National Natural Science Foundation of China under Grant U21B2041, 61825603, National Key R\&D Program of China 2020YFB2103902.}
\thanks{Yuyu Jia, Junyu Gao, Qiang Li, and Qi Wang are with the School of Artificial Intelligence, Optics and Electronics (iOPEN), Northwestern Polytechnical University, Xi’an 710072, Shaanxi, P. R. China.}
\thanks{Wei Huang is with the Data Science in Earth Observation, Technical University of Munich, Munich 80333, Germany.}
\thanks{E-mail: jyy2019@mail.nwpu.edu.cn, hw2hwei@gmail.com, gjy3035@gmail.com, crabwq@gmail.com, liqmges@gmail.com.}
\thanks{Qiang Li is the corresponding author.}
}

\maketitle

\begin{abstract}
Few-shot segmentation (FSS) for remote sensing (RS) imagery leverages supporting information from limited annotated samples to achieve query segmentation of novel classes. 
Previous efforts are dedicated to mining segmentation-guiding visual cues from a constrained set of support samples. 
However, they still struggle to address the pronounced intra-class differences in RS images, as sparse visual cues make it challenging to establish robust class-specific representations. 
In this paper, we propose a holistic semantic embedding (HSE) approach that effectively harnesses general semantic knowledge, \textit{i.e.,} class description (CD) embeddings.
Instead of the naive combination of CD embeddings and visual features for segmentation decoding, we investigate embedding the general semantic knowledge during the feature extraction stage.
Specifically, in HSE, a spatial dense interaction module allows the interaction of visual support features with CD embeddings along the spatial dimension via self-attention.
Furthermore, a global content modulation module efficiently augments the global information of the target category in both support and query features, thanks to the transformative fusion of visual features and CD embeddings.
These two components holistically synergize general CD embeddings and visual cues, constructing a robust class-specific representation.
Through extensive experiments on the standard FSS benchmark, the proposed HSE approach demonstrates superior performance compared to peer work, setting a new state-of-the-art.
\end{abstract}


\section{Introduction}
\label{Introduction}

\IEEEPARstart{S}{emantic} segmentation, entailing the pixel-level categorization of images, is essential for deciphering remote sensing imagery and acts as a foundational method across a wide range of practical applications\cite{zhang2017curriculum}, \cite{forster1985examination}, \cite{liu2023rotated}, \cite{ru2021land}, \cite{qiao2023weakly}, \cite{macioszek2021extracting}.
With the development of aerial and satellite imaging devices, the collection of large-scale remote sensing data becomes feasible, further driving advancements in deep learning for fully supervised remote sensing image segmentation.
However, deep learning-based methods \cite{Long_Shelhamer_2015}, \cite{Chen_Papandreo_2018} still struggle with the high cost of extensive annotation data.
Although researchers mitigate the stringent requirement for annotated data through semi-supervised \cite{chaudhuri2017multilabel}, \cite{10319694}, \cite{10466751}, \cite{10309935} and weakly supervised \cite{10097682}, \cite{9963946}, \cite{10189872}, \cite{10167684} perspectives, the generalization ability of models remains a bottleneck when confronted with entirely new categories.
As an effective approach to surmount these challenges, a plethora of few-shot segmentation (FSS) techniques \cite{Shaban_Bansal_Liu_Essa_Boots_2017}, \cite{Siam_Oreshkin_Jagersand_2019}, \cite{Yang_Liu_Li_Jiao_Ye_2020}, \cite{Nguyen_Todorovic_2019}, \cite{Wang_Liew_Zou_Zhou_Feng_2019}, \cite{Zhang_Lin_Liu_Yao_Shen_2019} have emerged.
Given a few annotated samples from a novel class, \textit{i.e.,} \textit{support}, FSS performs accurate segmentation on other samples within the same category, \textit{i.e.,} \textit{query}.

\begin{figure}[!]
\centering
\includegraphics[width=3.4in]{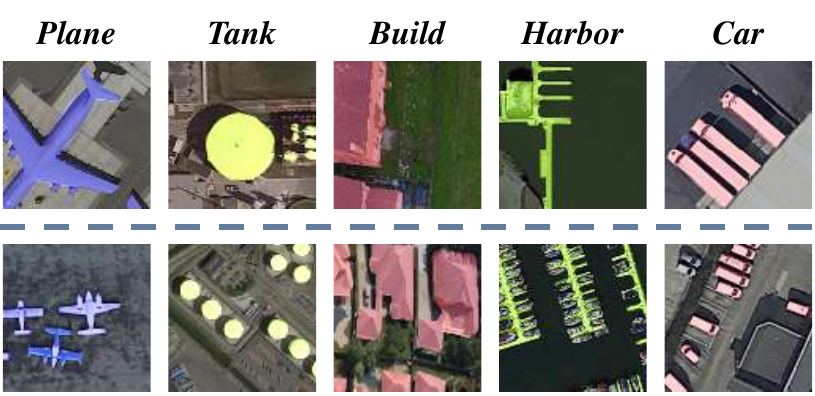}
\caption{Pronounced intra-class differences in remote sensing images.}
\label{fig_1}
\end{figure}

\begin{figure}[!]
\centering
\includegraphics[width=3.1in]{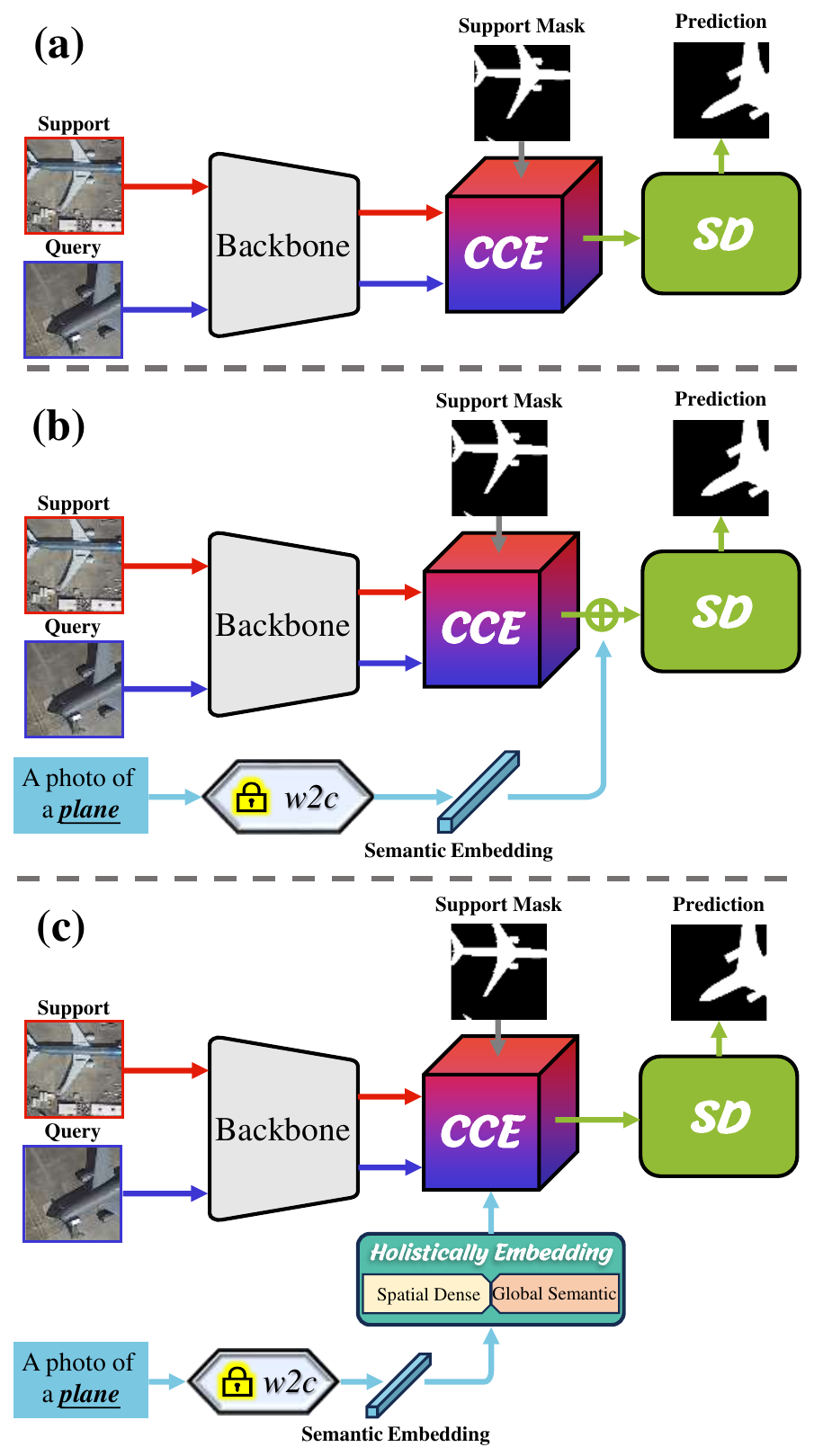}
\caption{Comparison between existing FSS methods and proposed HSE. (a) Many FSS algorithms adhere to a segmentation-guiding paradigm, primarily conducting research from two aspects: class-specific cues extraction (CCE) and segmentation decoder (SD). (b) Recent work, MIANet \cite{Yang_Chen_Feng_Huang}, introduces general semantic knowledge and combines it with visual support prototypes for segmentation guidance. (c) We further explore a holistic semantic embedding (HSE) approach that exploits the capabilities of general semantic knowledge through spatial dense interaction and global semantic modulation.}
\label{fig_2}
\end{figure}

As Fig. \ref{fig_2}(a) shows, FSS algorithms typically follow the segmentation-guided paradigm, exploring the extraction of class-specific visual cues from support samples and segmenting query samples through a decoder \cite{Zhang_Wei_Yang_Huang_2020}.
Some efforts are devoted to the segmentation decoding stage.
Specifically, prototypical learning methods \cite{Li_Jampani_Sevilla_2021}, \cite{Tian_Zhao_Shu_Yang_Li_Jia_2022}, \cite{Lang_Tu_Cheng_Han_2022} employ class prototypes as stand-ins for pixel-level correlations, indirectly guiding the segmentation of query images.
Affinity learning methods \cite{Shi_Wei_Zhang_Lu_Ning_Chen_Ma_Zheng}, \cite{Wang_Zhang_Hu_Yang_Cao_Zhen_2020}, \cite{Zhang_Kang_Wei_Yang_2021} directly transform pixel-level correlations into segmentation guidance information.
Another school of thought mines robust and comprehensive class-specific visual representations during the feature extraction stage, serving as segmentation-guiding cues.
For example, integrating multi-scale features \cite{Min_Kang_Cho_2021}, \cite{Wang_Suganuma_Okatani_2021}, \cite{Xie_Liu_Xiong_Shao_2021} leveraging deep semantic correlations \cite{Kang_Cho}, \cite{Hong_Cho_Nam_Lin_Kim_2022}, \cite{Liu_Jiao_Ye_2021} and incorporating multi-level support class information \cite{Moon_Sohn_Khan_Kapadia_2022}, \cite{Zhang_Xiao_Qin_2021}, \cite{Liu_Bao_Xie_Xiong_2022}, \textit{etc}.
Although showing promising results, these methods based on a single visual modality still have limitations.
A minuscule number of samples often falls short in furnishing robust visual representations for segmentation guidance, especially when dealing with remote sensing images characterized by pronounced intra-class differences, as illustrated in Fig. \ref{fig_1}.

To address the above issue, introducing general semantic knowledge, \textit{i.e.,} class description (CD) embeddings, can supplement some missing class information in visual representations since CD embeddings are responsible only for categories and are not influenced by individual sample differences.
Following this idea, AM3 \cite{Chen_Rostamzadeh_Oreshkin_2019} extracts textual prototypes from category descriptions and combines them with visual prototypes.
SP \cite{Chen_Si_Zhang_Wang_Wang_Tan_2023} modulates visual feature extraction using semantic information as prompts.
However, these methods are designed for classification tasks.
Recently, MIANet \cite{Yang_Chen_Feng_Huang} simply supplements CD embeddings with visual representation and achieves remarkable results (Fig. \ref{fig_2}(b)), which indicates its promising development space.

In this paper, we further explore a holistic semantic embedding (HSE) approach to harness the potential of general semantic knowledge.
First, CD embeddings for each class can be obtained by a powerful pre-trained language model, such as BERT \cite{Devlin_Chang_Lee_Toutanova_2019} and CLIP \cite{Radford_Kim_Hal_2021}.
Multi-level visual features are derived through the CNN backbone, \textit{i.e.,} ResNet50 \cite{He_Zhang_Ren_Sun_2016} or VGG16 \cite{Simonyan_Zisserman_2015}.
Inspired by \cite{Tian_Zhao_Shu_Yang_Li_Jia_2022}, we introduce the query prior mask via a prior generator to obtain coarse segmentation regions.
Subsequently, HSE is implemented as two serial complementary modules, \textit{i.e.,} spatial dense interaction (SDI) and global content modulation (GCM), to holistically embed general semantic knowledge into the visual feature extractor (Fig. \ref{fig_2}(c)).
Specifically, to facilitate the dense interaction \textit{on the spatial dimension} and inform the model to focus on class-specific spatial characteristics, SDI extends support features with CD embeddings and feeds them into an interactor implemented by a self-attention block.
Moreover, considering the global content matters the comprehensive understanding of the category scene, GCM converts CD embeddings and visual support prototypes into an enhancement coefficient through a modulator, to simultaneously enhance the support and query features \textit{on the channel dimension}.
By combining the two complementary modules, the proposed HSE approach effectively leverages the general semantic knowledge in class descriptions to mitigate intra-class individual differences in remote sensing imagery.
Through extensive experiments on the standard FSS benchmarks \textit{i.e.,} iSAID-$5^{i}$, HSE presents significant performance improvements with different types of pre-trained language encoders and few-shot settings, realizing the efficacy of general semantic knowledge for FSS tasks.

In summary, our contributions are three-fold:
\begin{enumerate}
     \item[1)] We propose a novel holistic semantic embedding approach to explore general semantic knowledge in class descriptions. To our best knowledge, we are the first to introduce textual modality information in few-shot remote sensing segmentation.

     \item[2)] To establish robust class-specific representations for segmentation guidance, we propose two complementary modules that holistically embed semantic knowledge into the visual feature extractor from both spatial dense and global content perspectives.
     
     \item[3)] Through comprehensive quantitative and qualitative analysis, we demonstrate the effectiveness and state-of-the-art performance of HSE on the FSS benchmark.
\end{enumerate}

\section{Related Works}
\label{Related_Works}

\subsection{Generic Semantic Segmentation}
\label{Generic_Semantic_Segmentation}

Semantic segmentation can provide pixel-level understanding for remote sensing images and has been extensively studied in recent years, driven by advancements in deep neural networks.
The introduction of Fully Convolutional Networks (FCNs) \cite{Long_Shelhamer_Darrell_2015} has pioneered the development of CNN-based methods.
Following this foundation, U-Net \cite{ronneberger2015u} introduces an encoder-decoder structure with skip connections, enriching the feature space information.
To enhance the model's multi-scale perception capability, a series of variants of atrous convolution \cite{Chen_Papandreou_2018}, \cite{Chen_Papandreou_Schroff_Adam_2017}, \cite{Chen_Zhu_Papandreou_Schroff_Adam_2018} and adaptive pooling \cite{Zhao_Shi_Qi_Wang_Jia_2017}, \cite{Zhao_Shi_Qi_Wang_Jia_2017}, \cite{Lian_Pang_Han_Pan_2021} techniques have emerged.
Later, transformer-based methods \cite{Strudel_Garcia_Laptev_Schmid_2021}, \cite{Xie_Wang_Yu_Anandkumar_Alvarez_Luo_2021}, \cite{Cao_Wang_Chen_Jiang_Zhang_Tian_Wang_2023}
aim to balance local and global spatial features, effectively improving semantic segmentation performance.

Drawing from successes in natural scenes, some researchers are dedicated to the task of remote sensing semantic segmentation.
Considering the broader scale variations of remote sensing objects, advanced methods predominantly employ attention mechanisms to extract global contextual information.
ResU-Net \cite{Li_Zheng_Duan_Su_Zhang_2022} designs a linear attention mechanism to approximate dot-product attention with significantly reduced computation costs, rendering the integration of attention mechanisms and CNN more flexible.
Given this, MANet \cite{Li_Zheng_Zhang_Duan_Su_Wang_Atkinson_2022}  integrates global dependencies in both channel and spatial dimensions.
SLCNet \cite{Yu_Ji_2023} supervises long-range correlations through category consistency information in the ground truth segmentation map.
CGGLNet \cite{10475326} leverages global contextual information to address unclear boundaries and incomplete structures.
However, the aforementioned methods are limited by the requirement for large-scale data training and struggle to adapt to new scenes.

\subsection{Few-Shot Learning}
\label{Few-Shot_Learning}
Few-shot learning endeavors to discern new categories with a limited set of labeled examples.
Recent investigations reveal two quintessential directions:
Optimization-based methods \cite{Finn_Abbeel_Levine_2017}, \cite{Nichol_Achiam_Schulman_2018}, \cite{Zintgraf_Shiarlis_Kurin_Hofmann_Whiteson_2018}, \cite{Rusu_Rao_Sygnowski_Vinyals_Pascanu_2018} learn a meta-learner to perform rapid adaption using a sparse set of training samples for novel categories.
Metric-based methodologies \cite{Vinyals_Blundell_Lillicrap_Kavukcuoglu_Wierstra_2016}, \cite{Snell_Swersky_Zemel_2017}, \cite{Fort_2018}, \cite{Doersch_Gupta_Zisserman_2020} cultivate a feature space wherein a suitable distance function is applied for the assessment of similarity.
More recently, large-scale language models (LLMs) and vision-language models (VLMs) have propelled advancements in few-shot learning. 
Some superior methods ingeniously leverage textual modal knowledge to compensate for the absence of visual features \cite{Chen_Si_Zhang_Wang_Wang_Tan_2023} or to directly enhance the few-shot classifiers \cite{Chen_Rostamzadeh_Oreshkin_2019}.

In the realm of remote sensing image processing, few-shot learning has likewise achieved significant progress.
For example, RS-MetaNet \cite{Li_Cui_Zhu_Chen_Zhu_Huang_Tao_2021} elevates the learning paradigm from individual samples to tasks through meta-training, thereby acquiring the ability to discern a metric space adept at classifying remote sensing scenes.
DLA-MatchNet \cite{Li_Han_Yao_Cheng_Guo_2021} employs the attention mechanism to discover discriminative regions.
HSL-MINet \cite{10182361} proposes a multiview-attention strategy designed to extract potentially shared information across various rotational perspectives of images.
MVP \cite{Zhu_Li_Qiu_Yang_Guan_Yi_2023} incorporates a parameter-efficient tuning method into the meta-learning framework, customizing it specifically for remote sensing images.
ICSFF \cite{10416681} amplifies feature representations for few-shot classification through a graph-structural feature fusion methodology.
Although these methods have achieved notable success in classification tasks, they lack the capability for pixel-level understanding of remote sensing images in few-shot scenarios.

\subsection{Few-Shot Segmentation}
\label{Few-Shot_Segmentation}
Few-shot semantic segmentation (FSS) aims to decode objects belonging to previously unseen categories with merely a handful of annotated samples.
OSLSM \cite{Shaban_Bansal_Liu_Essa_Boots_2017}, as a pioneering work in FSS, introduced a dual-branch segmentation paradigm.
Following this, various approaches can generally be categorized into two groups according to their segmentation decoding strategies:
prototypical learning methods \cite{Li_Jampani_Sevilla_2021}, \cite{Wang_Sun_Zhang_Zhang}, \cite{Zhang_Lin_Liu_Yao_Shen_2019} and affinity learning methods \cite{Shi_Wei_Zhang_Lu_Ning_Chen_Ma_Zheng}, \cite{Zhang_Kang_Wei_Yang_2021}, \cite{Zhang_Lin_Liu_Guo_Wu_Yao_2019}.
In the latest research, there is a state-of-the-art endeavor \cite{Yang_Chen_Feng_Huang} leverages the rich semantic knowledge in category descriptions to enhance the representation of support samples, achieving promising results.

Similarly, excellent research in few-shot remote sensing segmentation has increasingly come to the forefront.
For instance, the widely popular dataset iSAID-$5^{i}$ is introduced in \cite{Yao_Cao_Feng_Cheng_Han_2022}.
DMML-Net \cite{Wang_Wang_Sun_Wang_Fu_2022} constructs a deep feature pyramid comparison module and conceptualizes segmentation as a metric-based pixel classification task.
PCNet \cite{Lang_Wang_Tu_Han_Gong} progressively parses the entirety of the target region into local descriptors and designs base-class memories to distillate prototypes for novel classes.
HMRE \cite{10275071} enhances the target information within the dual-branch network structure through global semantic and spatially dense mutual reinforcement.
R2Net \cite{Lang_Cheng_Tu_Han_2023} implements dynamic global prototypes to mitigate inaccurate activations arising from intra-class differences.
MS2A2Net \cite{Li_Gong_Li_Zhang_Zhang_Wang_Wu} proposes an attention aggregation network to adaptively fuse multi-scale features and effectively model the foreground correlation of the dual-branch structure.
However, these methods are limited to extracting visual segmentation cues from a finite number of support samples and struggle with the significant intra-class differences in remote sensing images.  
This paper pioneers the incorporation of textual modal information from category descriptions in this task, addressing the shortfall of support information and providing fresh insights for this research.

\section{Methodology}

\subsection{Task Description}
\label{Task_Description}
Few-shot segmentation seeks to generalize a model's segmentation capability from a training dataset $\mathcal{D}_{train}$ to a completely novel category dataset $\mathcal{D}_{novel}$, corresponding to two disjoint sets of categories, $\mathcal{C}_{train}$ and $\mathcal{C}_{novel}$.
Consistent with prior studies\cite{Tian_Zhao_Shu_Yang_Li_Jia_2022}, we train the model episodically, where each episode includes a support set $\mathcal{S}$ and a query set $\mathcal{Q}$.
Given the $K$-shot setting, each support set comprises $K$ pairs of images $I_{s}$ and their corresponding binary masks $M_{s}$, where $\mathcal{S} = \left \{I_{s}^{k},M_{s}^{k} \right\}_{k=1}^{K}$.
Similarly, the query set is defined as $\mathcal{Q} = \left \{I_{q}, M_{q} \right\}$, where the mask $M_{q}$ of the query image is only available at the model training stage.
Our goal is to mine robust segmentation-guiding information from the support set to accurately parse the query images.
Note that we employ the $1$-shot setting to simplify the illustration of our approach.

\subsection{Method Overview}
\label{Method_Overview}
\begin{figure*}[!]
\centering
\includegraphics[width=7.2in]{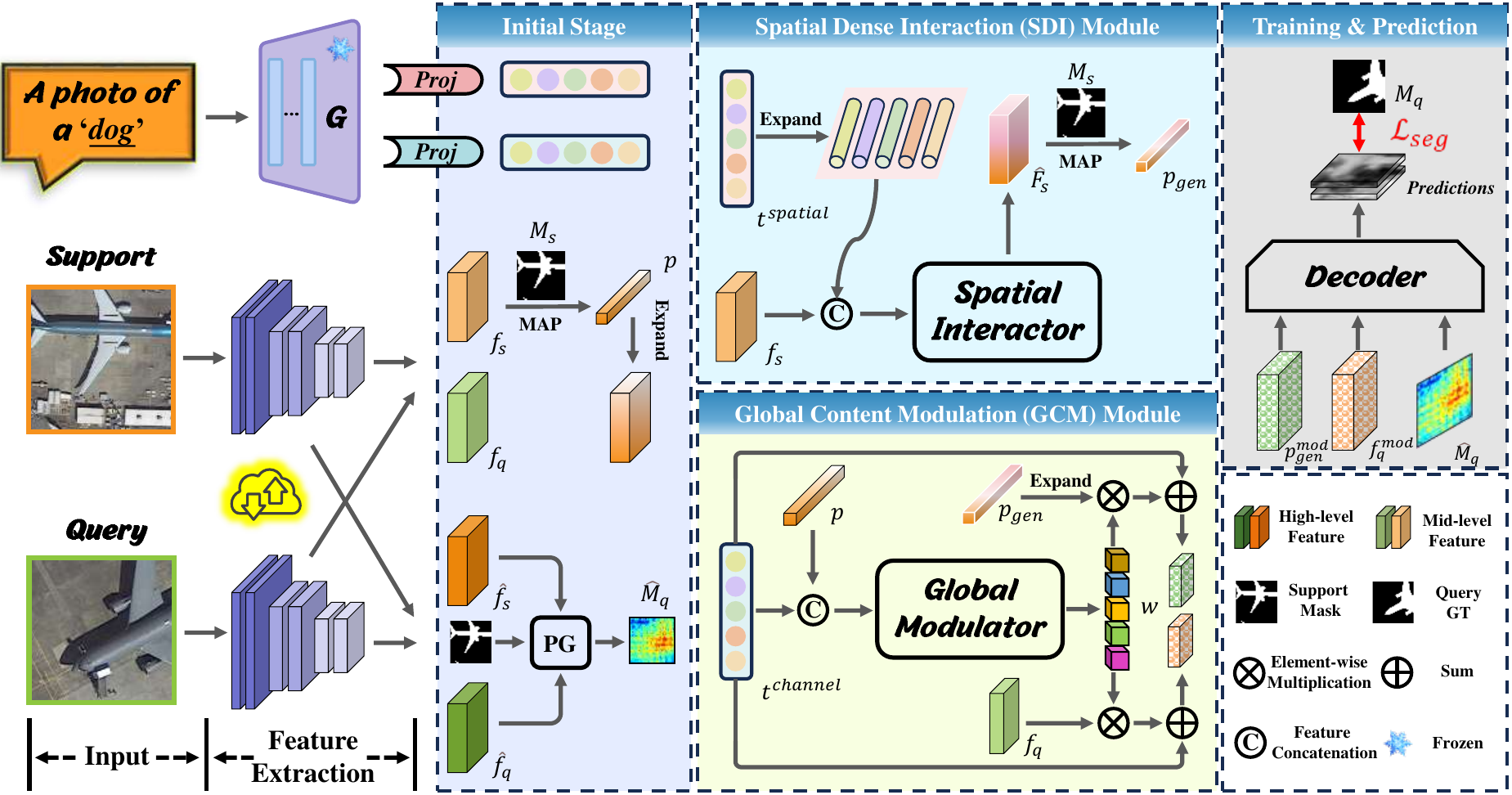}
\caption{Pipeline of the HSE method:
it first extracts mid- and high-level support and query features, query prior masks, and CD embeddings in the initial feature extraction stage.
To embed the general semantic knowledge from CD embeddings into visual cues and establish robust class-specific segmentation guidance, we design two sequential, complementary modules.
The SDI module facilitates the spatial dense interaction between general semantic knowledge and individual-specific visual features.
The GCM module enhances global content relevant to the target category in the support and query features through modulation coefficients.
Finally, along with the modulated query feature and query prior mask, the constructed robust class-specific representation assumes the role of segmentation guidance inputted into the decoder, yielding the query prediction mask.}
\label{fig_3}
\end{figure*}

As depicted in Fig. \ref{fig_3}, the proposed HSE approach encompasses an initial feature extraction stage \ref{Initial_Feature_Extraction_Stage} and two key complementary modules, the Spatial Dense Interaction (SDI) module \ref{Spatial_Dense_Interaction} and the Global Content Modulation (GCM) module \ref{Global_Semantic_Modulation}, that explore general semantic knowledge from category descriptions, constructing robust class-specific representations for segmentation guidance.
Specifically, given the support image $I_{s}$ and the query image $I_{q}$, two pre-trained backbones (ResNet50 \cite{He_Zhang_Ren_Sun_2016} or VGG16 \cite{Simonyan_Zisserman_2015}) with shared weights are adopted to extract mid- and high-level features.
We then obtain the class description (CD) embedding, support prototype, and query prior mask in the initial feature extraction stage. 
Subsequently, the holistic embedding of general semantic knowledge is executed in two sequential modules, \textit{i.e.,} SDI and GCM.
At last, we feed the robust class-specific guidance, the modulated query feature, and the query prior mask into a segmentation decoder, obtaining the prediction query mask.

\subsection{Initial feature Extraction Stage}
\label{Initial_Feature_Extraction_Stage}

The core issue in few-shot segmentation is how to transform the provided support image and its mask into class-specific segmentation guidance information.
Among a plethora of studies, the predominant strategy is to compress the foreground region features of support images into prototypes, which are then used to guide the segmentation process of query images.
To elaborate, background information in the middle-level support features is masked out, leaving only foreground information, which is then concentrated in the prototypes through average pooling.
Formally, given the mid-level support feature $f_{s} \in \mathbb{R}^{C\times H\times W}$ and the corresponding support mask $M_{s} \in \mathbb{R}^{H \times W}$, where $C, H, W$ represent the channel, height, and weight, the prototype $p \in \mathbb{R}^{C \times 1 \times 1}$ can be expressed as:
\begin{equation}
\label{eqt_1}
p = \mathcal{A}(f_{s}\otimes \mathcal{R}(M_{s})),
\end{equation}
where $\mathcal{A}(\cdot)$ represents the average pooling operation, $\otimes$ denotes the Hadamard product, and $\mathcal{R}$ is a spatial scaling function.

Regarding the utilization of high-level features, the esteemed PFENet \cite{Tian_Zhao_Shu_Yang_Li_Jia_2022} transforms them into a class-agnostic prior mask to roughly indicate the target area's location in the query image.
The prior mask $\hat{M}_{q} \in \mathbb{R}^{H \times W}$ is derived from the maximal element-wise correlation responses between the high-level features of the query and support. 
We abstract this process as a prior mask generator $\mathcal{PG}$, which takes support masks $M_{s}$, high-level features $\hat{f}_{s}$ and $\hat{f}_{q}$ as inputs. 
This can be represented as:
\begin{equation}
\label{eqt_2}
\hat{M}_{q} = \mathcal{PG}(\hat{f}_{s}, \hat{f}_{q}, M_{s}).
\end{equation}

Moreover, with the pre-trained language model $\mathcal{G}(\cdot)$, we transform the category \textit{name} into the CD embedding $t = \mathcal{G}(\textit{name}) \in \mathbb{R}^{C_{t} \times 1 \times 1}$, where $C_{t}$ is dimension of the embedding space.

\subsection{Spatial Dense Interaction}
\label{Spatial_Dense_Interaction}
Existing studies revolve around the two-level segmentation guidance cues, \textit{i.e.,} $p$ and $\hat{M}_{q}$, which are finally passed to the decoder along with the mid-level query features for segmenting the target region.
However, they still struggle to break through the limitations imposed by the single visual modality, especially in the face of remote sensing images that exhibit particularly substantial intra-class differences.
Essentially, the segmentation guidance cues $p$ and $\hat{M}_{q}$ are both extracted from sparse, individual-specific visual features.
They lack sufficient general category information to construct robust class-specific representations, rendering it arduous to efficiently encode support-query image pairs with visual disparities.
In contrast, individual-agnostic CD embeddings encompass general semantic knowledge.
Building on this concept, we propose a spatial dense interaction (SDI) module to embed the semantic knowledge from the spatial dimension to supplement missing general information in visual cues.

\begin{figure}[!]
\centering
\includegraphics[width=3.45in]{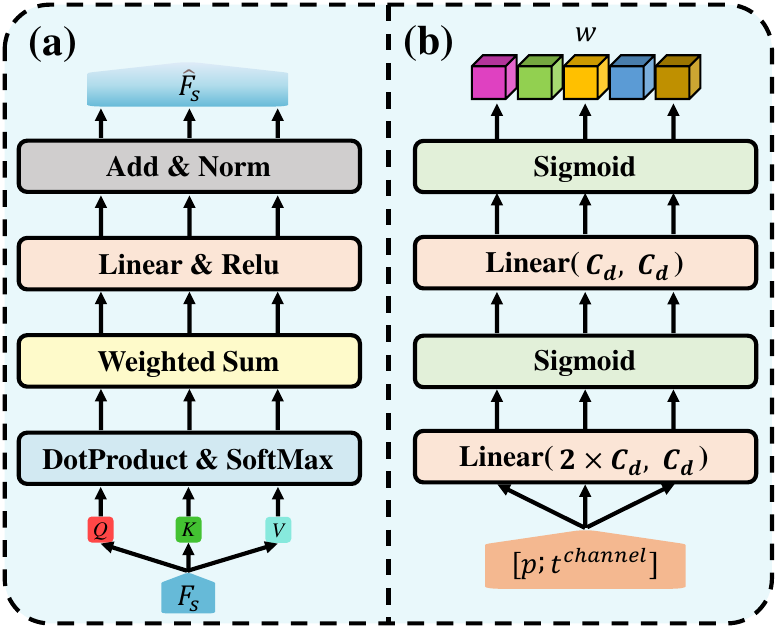}
\caption{Structure of the SDI (a) and GCM (b) modules.}
\label{fig_4}
\end{figure}

Inspired by prompt learning methods \cite{Petroni_Rocktäschel_Riedel_Lewis_Bakhtin_Wu_Miller_2019}, \cite{Brown_Mann_2020}, we concatenate CD embeddings with visual feature maps and further promote the interaction between the two through a self-attention interactor (Fig. \ref{fig_4} (a)), achieving semantic knowledge embedding.
Given the mid-level support feature $f_{s}$ of a certain category and its CD embedding $t \in \mathbb{R}^{C_{t} \times 1 \times 1}$, we obtain a new feature map $F_{s} \in \mathbb{R}^{C \times (HW+W)}$ by extending $f_{s}$ with the projected CD embedding:
\begin{equation}
\label{eqt_3}
F_{s} = [f_{s}; \textit{repeat}(t^{spatial})],
\end{equation}
where $t^{spatial} = \mathcal{P}_{spatial}(t)$, $\mathcal{P}_{spatial}(\cdot)$ is a projector that maintains the channel dimension to be the same as the visual feature, $[\cdot;\cdot]$ denotes the feature concatenation operation, and $\textit{repeat}(\cdot)$ expands the dimension through repetition operations.

Then, the extended feature map $F_{s}$ is fed into the \textit{interactor}, in which a self-attention block allows the flow of general semantic information to visual features.
Formally, three input vectors \textbf{\textit{q}}, \textbf{\textit{k}}, \textbf{\textit{v}} of the attention block have the same magnitude as $F_{s}$, the output feature map $\hat{F}_{s} \in \mathbb{R}^{C \times H \times W}$ can be calculated by:
\begin{equation}
\label{eqt_4}
\hat{F}_{s} = Resize(interactor(F_{s}, F_{s}, F_{s})[:, :HW]),
\end{equation}
where $\textit{Resize}(\cdot)$ restores the spatial dimension $H \times W$ of the feature map.

Finally, we further compute the general support prototype $p_{gen}$ supplemented with the semantic knowledge as:
\begin{equation}
\label{eqt_5}
p_{gen} = \mathcal{A}(\hat{F}_{s} \otimes \mathcal{R}(M_{s})).
\end{equation}

\subsection{Global Content Modulation}
\label{Global_Semantic_Modulation}
Features from channels often contain rich global semantic content, which is closely linked to the contextual understanding of the target scene.
Therefore, besides embedding general semantic knowledge, \textit{i.e.,} CD embeddings, within the spatial dimension, it is equally crucial to modulate visual features on a channel-by-channel basis.

Given the visual support prototype $p \in \mathbb{R}^{C \times 1 \times 1 \times 1}$ of a certain category and its CD embedding $t \in \mathbb{R}^{C_{t} \times 1 \times 1}$, we first convert the two to an enhancement coefficient by a modulator:
\begin{equation}
\label{eqt_6}
w = \textit{modulator}[p; t^{channel}],
\end{equation}
where $w \in \mathbb{R}^{C \times 1 \times 1}$, $t^{channel} = \mathcal{P}_{channel}(t)$, $\mathcal{P}_{channel}(\cdot)$ is a projector that keeps the channel dimension consistent with the visual prototype, and the structure of the $\textit{modulator}$ is illustrated in Fig. \ref{fig_4} (b).

Then, the general prototype and query feature can be modulated by the enhancement coefficient:
\begin{equation}
\label{eqt_7}
\left\{\begin{matrix}
 p_{gen}^{mod} = p_{gen} \otimes w + t^{channel}
\\
 f_{q}^{mod} = f_{q} \otimes w + t^{channel}
\end{matrix}\right..
\end{equation}
Through such embedding of channel-level general semantic knowledge, the global information specific to the target categories contained within the support and query branches is accentuated, and the interference caused by intra-class individual differences is effectively suppressed.

\subsection{Prediction and Training Loss}
\label{Prediction_and_Training_Loss}
We employ the commonly used encoder-decoder structure in segmentation tasks \cite{Kang_Cho}, \cite{Hong_Cho_Nam_Lin_Kim_2022}.
Having obtained the modulated general prototype $p_{gen}^{mod}$ and modulated query feature $f_{q}^{mod}$ from the aforementioned encoder (Section  \ref{Initial_Feature_Extraction_Stage}-\ref{Global_Semantic_Modulation}), we aggregate them into the segmentation decoder $\mathcal{D}(\cdot)$ to attain the final prediction $Y \in \mathbb{R}^{H \times W}$:
\begin{equation}
\label{eqt_8}
Y = \mathcal{D}([p_{gen}^{mod};f_{q}^{mod}]).
\end{equation}
As $p_{gen}^{mod}$ incorporates a holistic embedding of general semantic knowledge across both spatial dense and global content perspectives, it is equipped to provide robust segmentation guidance for the query image.
The segmentation loss is computed utilizing the standard binary cross-entropy (BCE) function:
\begin{equation}
\label{eqt_9}
\mathcal{L} = \mathcal{D}([p_{gen}^{mod};f_{q}^{mod}]).
\end{equation}

\subsection{Extending to K-Shot Setting}
\label{Extending_to_K-Shot_Setting}
In the $K$-shot scenario, $K$ support pairs $\left \{I_{s}^{k},M_{s}^{k} \right\}_{k=1}^{K}$ are available to provide visual segmentation guidance.
Our method requires only minimal adjustments to address the $K$-shot setting.
In the initial feature extraction stage, the $K$ visual prototypes and query prior masks computed by \textit{Eq.} \ref{eqt_1} and \ref{eqt_2} are averaged.
In addition, $K$ support samples spatial densely interact with their respective CD embeddings, yielding $\left \{p_{gen}^{k} \right\}_{k=1}^{K}$. 
The final $p_{gen}$ is obtained by averaging these values as follows:
\begin{equation}
\label{eqt_10}
p_{gen} = \frac{1}{K}\sum_{k=1}^{K}p_{gen}^{k}.
\end{equation}

\section{Experiments}
To ascertain the efficacy of the proposed HSE method, both 1-shot and 5-shot settings are adhered to, conducting ample quantitative and qualitative experiments on the publicly available FSS dataset.
First, we delineate the FSS benchmark and the corresponding evaluation metric.
Then, the implementation specifics of HSE are elucidated to facilitate realization.
Following this, we present comparisons between the HSE method and other leading counterparts, analyzing the segmentation results in a comprehensive manner.
Ultimately, a series of ablation studies are executed to demonstrate the impact of each component within the HSE framework.

\begin{table*}[!]
  \centering
  \caption{Category Distribution and Splitting of iSAID-$5^{i}$ Dataset.}
  \setlength{\tabcolsep}{0.75mm}{
    \begin{tabular}{c|ccccc|ccccc|ccccc}
    \hline
    Split ID & \multicolumn{5}{c|}{iSAID-$5^{0}$}         & \multicolumn{5}{c|}{iSAID-$5^{1}$}         & \multicolumn{5}{c}{iSAID-$5^{2}$} \bigstrut\\
    \hline
    Category ID & C1    & C2    & C3    & C4    & C5    & C6    & C7    & C8    & C9    & C10   & C11   & C12   & C13   & C14   & C15 \bigstrut[t]\\
    Category name & ship  & \makecell[c]{storage \\ tank}  & \makecell[c]{baseball \\ diamond}  & \makecell[c]{tennis \\ court}  & \makecell[c]{basketball \\ court} & \makecell[c]{ground track \\ field}  & bridge & \makecell[c]{large \\ vehicle} & \makecell[c]{small \\ vehicle} & helicopter & \makecell[c]{swimming \\ pool} & roundabout & \makecell[c]{soccer ball \\ field} & plane & harbor \\
    \makecell[c]{Num of imgs \\ in train set}     & 3820  & 902   & 495   & 2485  & 598   & 1405  & 224   & 2953  & 2592  & 69    & 147   & 203   & 2000  & 1864  & 3358 \\
    \makecell[c]{Num of imgs \\ in test set}     & 1392  & 269   & 221   & 767   & 160   & 517   & 91    & 789   & 781   & 21    & 45    & 45    & 682   & 990   & 1361 \bigstrut[b]\\
    \hline
    \end{tabular}}%
  \label{tab_1}%
\end{table*}%

\subsection{Setup}
\label{Setup}

\subsubsection{Dataset}
Despite the recent emergence of a few FSS datasets for remote sensing images, they still lack a unified partitioning standard, making it difficult to fairly compare the performance of different algorithms.
In response, we adopt the iSAID-$5^{i}$ dataset \cite{Yao_Cao_Feng_Cheng_Han_2022} widely used by classical algorithms.
It encompasses 15 categories, with 18076 images designated for training and 6363 images for testing.
Each sample is dimensioned at 3×256×256.
The specific distribution of categories and their partitioning are illustrated in Table \ref{tab_1}.

\subsubsection{Evaluation Metric}
Building upon prior research in the domain \cite{Yao_Cao_Feng_Cheng_Han_2022}, \cite{Lang_Wang_Tu_Han_Gong}, we utilize the mean Intersection over Union (mIoU) as the evaluation metric.
Formally, let $N$ be the number of test (unseen) categories in each target fold, and the $IoU$ of category c is represented by $IoU_{c}$, the $mIoU$ can be defined as:
\begin{equation}
\label{eqt_11}
mIoU = \frac{1}{N}\sum_{c=1}^{N}IoU_{c}.
\end{equation}

\subsection{Implementation Details}
\label{Implementation_Details}
HSE is constructed using the PyTorch \cite{paszke2019pytorch} framework, and all experiments are conducted on a single NVIDIA GeForce RTX 3090 GPU.
We leverage ResNet50 \cite{He_Zhang_Ren_Sun_2016} and VGG16 \cite{Simonyan_Zisserman_2015} pre-trained on ImageNet \cite{krizhevsky2012imagenet} as feature extractors.
To preserve the generalizability of the feature extractor, we follow \cite{Tian_Zhao_Shu_Yang_Li_Jia_2022} to freeze its parameters.
To derive general semantic knowledge from class name embeddings, three types of language models pre-trained on large-scale corpora are utilized, \textit{i.e.,} CLIP \cite{Radford_Kim_Hal_2021}, SBERT \cite{Reimers_Gurevych_2019}, and GloVe \cite{Pennington_Socher_Manning_2014}.
We augment the category name with a text template as input: \textit{A photo of a `category name'}.

During meta-training, we employ the stochastic gradient descent (SGD) optimizer with an initial learning rate of 0.005, a momentum of 0.9, and a weight decay of 0.0001.
The batch size is set at 16, and the training epoch is fixed at 100.

During the meta-testing phase, we randomly sample 2000 episodes for model evaluation, with the final results obtained by averaging the outcomes of experiments influenced by five different random seeds.

\subsection{Comparison With State-of-the-Art}
\label{Comparison_With_State-of-the-Art}
In this section, we compare the proposed HSE method against other advanced FSS algorithms with the available source codes or public experimental results on the iSAID-$5^{i}$ dataset.
Under standardized experimental conditions, the quantitative and qualitative comparison results are presented to rigorously validate the superiority of HSE.

\subsubsection{Quantitative Analysis}
The performance comparison of our proposed HSE against other exemplary FSS algorithms is depicted in Table \ref{tab_2}, mIoU is leveraged as the evaluation metric under $1$-shot and $5$-shot settings. 
The best performance values are highlighted in bold, and the second-best results are marked with an underline. 
Overall, the proposed HSE method achieves the best segmentation performance across various backbones and support sample sizes.
Specifically, with the ResNet50 backbone, HSE achieves the best performance in every split, enhancing the average mIoU by 2.73\% under $1$-shot settings.
For the $5$-shot setting, HSE exhibits a distinct advantage, achieving an average performance of 43.15\%, representing an improvement of 2.09\%.
Simultaneously, employing the VGG16 backbone, HSE achieved the best performance for each split under the $1$-shot setting, improving the average performance by 0.97\%.
Similar to the results for the $5$-shot setting, HSE led in both split1 and split2, surpassing the HMRE method by an average mIoU of 1.51\%.
In terms of comprehensive analysis, the proposed HSE method demonstrates significant advantages in FSS tasks, particularly excelling in 1-shot scenarios. This effectively validates that incorporating general semantic knowledge is beneficial for establishing robust class-specific segmentation guidance.

To specifically analyze the model's performance across various categories, we detail the segmentation results under the $1$-shot setting in Table. \ref{tab_2}.
The proposed HSE achieves excellent performance across the majority of categories.
While some methods get better segmentation results in specific categories, their performance is not consistently stable.
For example, the BAM algorithm performs well in segmenting tennis court (C4), but it struggles with smaller-sized targets such as harbor (C15).
The DML method also excels at parsing tennis court (C4), yet it encounters difficulties with the small vehicle category (C9).
In contrast, our method still maintains good segmentation capabilities for challenging categories, exhibiting a more stable overall performance. 
This indicates that HSE has a stronger adaptability to complex scenarios, benefiting from the embedding of general semantic knowledge that compensates for the lack of sparse visual representation information.

\subsubsection{Qualitative Analysis}

\begin{figure*}[!]
\centering
\includegraphics[width=6.0in]{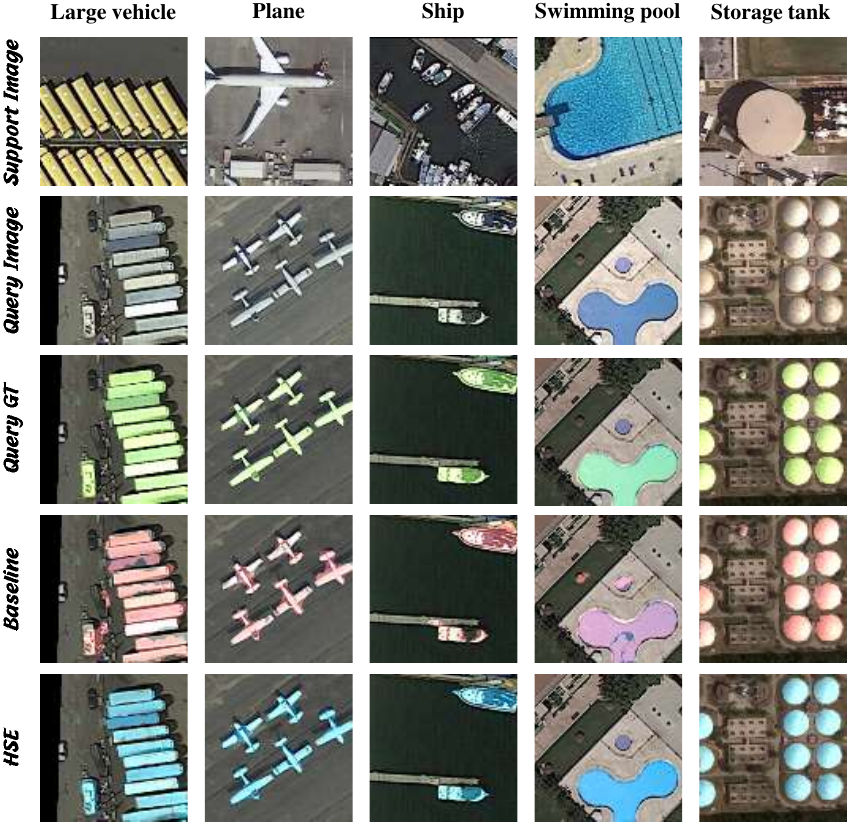}
\caption{Quantitative comparison of the segmentation effects of the HSE and baseline methods on iSAID-$5^{i}$ dataset under the $1$-shot setting.}
\label{fig_6}
\end{figure*}

To visually analyze the efficacy of the proposed HSE, Fig. \ref{fig_6} illustrates quantitative segmentation results in the $1$-shot scenario, where the baseline approach represents HSE with the GCM and SDI components removed.
Compared with the baseline method, HSE accurately activates target scene areas, generating prediction masks with more finely detailed edges.
Moreover, HSE effectively reduces false activations in areas of the query image that are similar to the target category.
Particularly, for the ``ship" and ``swimming pool" categories, the baseline method misclassifies similar class regions in the background, while HSE effectively suppresses the aforementioned interference.

\begin{table*}[!]
  \centering
  \caption{Performance Comparison on iSAID-$5^{i}$ Dataset under $1$-Shot and $5$-Shot Settings. Bolded Values Indicate the Best Performance, while the Second-Best Performances are Underlined.}
  \setlength{\tabcolsep}{4.5mm}{
    \begin{tabular}{l|rrrr|rrrr}
    \hline
    \multicolumn{1}{c|}{\multirow{2}[4]{*}{Method}} & \multicolumn{4}{c|}{1-shot}   & \multicolumn{4}{c}{5-shot} \bigstrut\\
\cline{2-9}          & \multicolumn{1}{l}{Split0} & \multicolumn{1}{l}{Split1} & \multicolumn{1}{l}{Split2} & \multicolumn{1}{l|}{Mean} & \multicolumn{1}{l}{Split0} & \multicolumn{1}{l}{Split1} & \multicolumn{1}{l}{Split2} & \multicolumn{1}{l}{Mean} \bigstrut\\
    \hline
    \multicolumn{9}{c}{ResNet50} \bigstrut\\
    \hline
    PANet \cite{Wang_Liew_Zou_Zhou_Feng_2019} & 27.56 & 17.23 & 24.60 & 23.13 & 36.54 & 16.05 & 26.22 & 26.27 \bigstrut[t]\\
    CANet \cite{Zhang_Lin_Liu_Yao_Shen_2019} & 25.51 & 13.50 & 24.45 & 21.15 & 29.32 & 21.85 & 26.91 & 26.03 \\
    SCL \cite{Zhang_Xiao_Qin_2021} & 34.78 & 22.77 & 31.20 & 29.58 & 41.29 & 25.73 & 37.70 & 34.91 \\
    PFENet \cite{Tian_Zhao_Shu_Yang_Li_Jia_2022} & 35.84 & 23.35 & 27.20 & 28.80 & 42.42 & 25.34 & 33.00 & 33.59 \\
    NERTNet \cite{Liu_Liu_Cao_Yao_Han_Shao} & 34.93 & 23.95 & 28.56 & 29.15 & 44.83 & 26.73 & 37.19 & 36.25 \\
    DCP \cite{Lang_Tu_Cheng_Han_2022} & 37.83 & 22.86 & 28.92 & 29.87 & 41.52 & 28.18 & 33.43 & 34.38 \\
    BAM \cite{Lang_Cheng_Tu_Han} & 39.43 & 21.69 & 28.64 & 29.92 & 43.29 & 27.92 & 38.62 & 36.61 \\
    DMML \cite{Wang_Wang_Sun_Wang_Fu_2022} & 28.45 & 21.02 & 23.46 & 24.31 & 30.61 & 23.85 & 24.08 & 26.18 \\
    DML \cite{Jiang_Zhou_Li_2022} & 32.96 & 18.98 & 26.27 & 26.07 & 33.58 & 22.05 & 29.77 & 28.47 \\
    SDM \cite{Xie_Liu_Xiong_Shao_2021} & 27.96 & 21.99 & 27.82 & 25.92 & 28.50 & 25.23 & 31.07 & 28.27 \\
    SD-AANet \cite{Zhao_Liu_Lyu_Wang_Yang_2021} & 30.52 & 29.31 & 18.46 & 26.10 & 28.01 & 21.38 & 16.79 & 22.06 \\
    TBPN \cite{P_Verma} & 29.33 & 16.84 & 25.47 & 23.88 & 30.98 & 20.42 & 28.07 & 26.49 \\
    HMRE \cite{10182361} & 40.88 & \ul{32.88} & \ul{37.22} & \ul{36.99} & 42.41 & \ul{34.69} & \textbf{46.07} & \ul{41.06} \\
    R$^{2}$Net \cite{Lang_Cheng_Tu_Han_2023} & \ul{41.22} & 21.64 & 35.28 & 32.71 & \ul{46.45} & 25.80 & 39.84 & 37.36 \\
    PCNet \cite{Lang_Wang_Tu_Han_Gong} & 40.24 & 24.64 & 31.31 & 32.06 & 45.31 & 28.19 & 37.36 & 36.95 \\
    \cellcolor[gray]{.8} HSE(\textit{ours}) & \cellcolor[gray]{.8} \textbf{44.03} & \cellcolor[gray]{.8} \textbf{35.87} & \cellcolor[gray]{.8} \textbf{39.26} & \cellcolor[gray]{.8} \textbf{39.72} & \cellcolor[gray]{.8} \textbf{47.03} & \cellcolor[gray]{.8} \textbf{36.75} & \cellcolor[gray]{.8} \ul{45.66} & \cellcolor[gray]{.8} \textbf{43.15} \bigstrut[b]\\
    \hline
    \multicolumn{9}{c}{VGG16} \bigstrut\\
    \hline
    PANet \cite{Wang_Liew_Zou_Zhou_Feng_2019} & 26.86 & 14.56 & 20.69 & 20.70 & 30.89 & 16.63 & 24.05 & 23.86 \bigstrut[t]\\
    CANet \cite{Zhang_Lin_Liu_Yao_Shen_2019} & 13.91 & 12.94 & 13.67 & 13.51 & 17.32 & 15.07 & 18.23 & 16.87 \\
    SCL \cite{Zhang_Xiao_Qin_2021} & 25.75 & 18.57 & 22.24 & 22.19 & 35.77 & 24.92 & 32.70 & 31.13 \\
    PFENet \cite{Tian_Zhao_Shu_Yang_Li_Jia_2022} & 28.52 & 17.05 & 18.94 & 21.50 & 37.59 & 23.22 & 30.45 & 30.42 \\
    NERTNet \cite{Liu_Liu_Cao_Yao_Han_Shao} & 25.78 & 20.01 & 19.88 & 21.89 & 38.43 & 24.21 & 28.99 & 30.54 \\
    DCP \cite{Lang_Tu_Cheng_Han_2022} & 28.17 & 16.52 & 22.49 & 22.39 & 39.65 & 22.68 & 29.93 & 30.75 \\
    BAM \cite{Lang_Cheng_Tu_Han} & 33.93 & 16.88 & 21.47 & 24.09 & 38.46 & 22.76 & 28.81 & 30.01 \\
    DMML \cite{Wang_Wang_Sun_Wang_Fu_2022} & 24.41 & 18.58 & 19.46 & 20.82 & 28.97 & 21.02 & 22.78 & 24.26 \\
    DML \cite{Jiang_Zhou_Li_2022} & 30.99 & 14.60 & 19.05 & 21.55 & 34.03 & 16.38 & 26.32 & 25.58 \\
    SDM \cite{Xie_Liu_Xiong_Shao_2021} & 24.52 & 16.31 & 21.01 & 20.61 & 26.73 & 19.97 & 26.10 & 24.27 \\
    SD-AANet \cite{Zhao_Liu_Lyu_Wang_Yang_2021} & 28.01 & 21.38 & 16.79 & 22.06 & 34.89 & 25.76 & 16.81 & 25.82 \\
    TBPN \cite{P_Verma} & 27.86 & 12.32 & 18.16 & 19.45 & 32.79 & 16.28 & 24.27 & 24.45 \\
    HMRE \cite{10182361} & \ul{36.48} & \ul{28.28} & \ul{35.36} & \ul{33.37} & 37.33 & \ul{29.28} & \ul{43.36} & \ul{36.66} \\
    R$^{2}$Net \cite{Lang_Cheng_Tu_Han_2023} & 35.27 & 19.93 & 24.63 & 26.61 & \textbf{42.06} & 23.52 & 30.06 & 31.88 \\
    PCNet \cite{Lang_Wang_Tu_Han_Gong} & 32.48 & 19.88 & 24.56 & 25.64 & \ul{41.09} & 21.98 & 34.14 & 32.40 \\
    \cellcolor[gray]{.8} HSE(\textit{ours}) & \cellcolor[gray]{.8} \textbf{37.18} & \cellcolor[gray]{.8} \textbf{29.79} & \cellcolor[gray]{.8} \textbf{36.04} & \cellcolor[gray]{.8} \textbf{34.34} & \cellcolor[gray]{.8} 40.87 & \cellcolor[gray]{.8} \textbf{30.20} & \cellcolor[gray]{.8} \textbf{43.44} & \cellcolor[gray]{.8} \textbf{38.17} \bigstrut[b]\\
    \hline
    \end{tabular}}%
  \label{tab_2}%
\end{table*}%

\begin{table*}[!]
  \centering
  \caption{Performance Comparison across Specific Categories on iSAID-$5^{i}$ Dataset under the $1$-Shot Setting.}
  \setlength{\tabcolsep}{1mm}{
    \begin{tabular}{l|rrrrrrrrrrrrrrr}
    \hline
    \multicolumn{1}{l}{Category ID} & \multicolumn{1}{l}{C1} & \multicolumn{1}{l}{C2} & \multicolumn{1}{l}{C3} & \multicolumn{1}{l}{C4} & \multicolumn{1}{l}{C5} & \multicolumn{1}{l}{C6} & \multicolumn{1}{l}{C7} & \multicolumn{1}{l}{C8} & \multicolumn{1}{l}{C9} & \multicolumn{1}{l}{C10} & \multicolumn{1}{l}{C11} & \multicolumn{1}{l}{C12} & \multicolumn{1}{l}{C13} & \multicolumn{1}{l}{C14} & \multicolumn{1}{l}{C15} \bigstrut\\
    \hline
    \multicolumn{16}{c}{ResNet50} \bigstrut\\
    \hline
    PANet \cite{Wang_Liew_Zou_Zhou_Feng_2019} & 21.81 & 36.31 & 23.01 & 42.06 & 14.59 & 12.11 & 17.44 & 22.70 & 12.27 & 21.60 & \ul{30.29} & 24.62 & 26.79 & 25.54 & 15.79 \bigstrut[t]\\
    CANet \cite{Zhang_Lin_Liu_Yao_Shen_2019} & 39.57 & 18.54 & 18.46 & 33.63 & 17.34 & 9.78 & 5.49 & 22.15 & 5.17 & 24.89 & 9.96 & 36.50 & 19.12 & 38.82 & 17.85 \\
    SCL \cite{Zhang_Xiao_Qin_2021} & 37.61 & 33.63 & 26.68 & 54.75 & 21.22 & 22.60 & 24.40 & 30.22 & 6.71 & 29.93 & \textbf{33.00} & 44.68 & 18.25 & \textbf{44.63} & 15.46 \\
    PFENet \cite{Tian_Zhao_Shu_Yang_Li_Jia_2022} & 39.02 & 45.63 & 20.86 & 49.96 & 23.72 & 21.00 & 24.76 & 31.59 & 6.98 & 32.42 & 13.34 & 47.64 & 30.65 & 32.82 & 11.54 \\
    NERTNet \cite{Liu_Liu_Cao_Yao_Han_Shao} & 33.59 & 42.83 & 22.30 & 49.35 & 21.91 & 21.62 & 28.82 & 25.64 & 9.35 & \ul{34.30} & 23.91 & 38.67 & 25.63 & 40.84 & 13.74 \\
    DCP \cite{Lang_Tu_Cheng_Han_2022} & 37.42 & 42.44 & 35.16 & \ul{56.55} & 17.58 & 21.66 & 19.57 & 32.97 & 10.60 & 29.50 & 24.02 & 35.34 & 28.44 & 39.80 & 17.02 \\
    BAM \cite{Lang_Cheng_Tu_Han} & 36.34 & 39.76 & 38.23 & \textbf{58.13} & 24.71 & 18.25 & 12.68 & \ul{35.91} & 11.42 & 30.21 & 28.98 & 40.74 & 29.43 & 33.25 & 10.79 \\
    DMML \cite{Wang_Wang_Sun_Wang_Fu_2022} & 40.14 & 40.18 & 21.31 & 27.02 & 13.60 & 15.56 & 15.19 & 26.05 & 13.84 & \textbf{34.44} & 11.26 & 17.57 & 23.27 & 39.11 & \ul{26.12} \\
    DML \cite{Jiang_Zhou_Li_2022} & 35.13 & 42.10 & 30.49 & 41.79 & 15.31 & 13.25 & 16.87 & 24.70 & 14.62 & 25.45 & 10.24 & 35.49 & 25.35 & 41.69 & 18.57 \\
    SDM \cite{Xie_Liu_Xiong_Shao_2021} & 41.77 & 35.50 & 21.41 & 20.81 & 20.29 & 15.60 & 25.60 & 28.66 & 13.29 & 26.79 & 13.61 & 32.35 & 24.59 & 42.79 & 25.75 \\
    SD-AANet \cite{Zhao_Liu_Lyu_Wang_Yang_2021} & 25.21 & 39.85 & \textbf{43.60} & 21.69 & 22.27 & \textbf{35.18} & \textbf{47.28} & 23.89 & 25.47 & 14.75 & 9.46 & 27.38 & 20.51 & 12.50 & 22.45 \\
    TBPN \cite{P_Verma} & 25.36 & 41.28 & 30.67 & 32.88 & 16.48 & 13.48 & 9.74 & 27.88 & 12.52 & 20.56 & 11.12 & 34.31 & 23.57 & 40.36 & 17.98 \\
    HMRE \cite{10182361} & 41.68 & 41.15 & 40.01 & 41.04 & \ul{40.52} & 27.36 & 34.66 & 35.44 & \textbf{39.34} & 27.58 & 24.56 & \ul{49.34} & \textbf{46.82} & \ul{4.27} & 21.11 \\
    R$^{2}$Net \cite{Lang_Cheng_Tu_Han_2023} & \textbf{46.87} & \textbf{49.06} & 30.70 & 52.86 & 26.62 & 24.31 & 17.25 & 31.25 & 13.67 & 21.73 & 24.88 & 46.07 & 42.29 & 42.07 & 21.08 \\
    \cellcolor[gray]{.8} HSE(\textit{ours}) & \cellcolor[gray]{.8} \ul{45.67} & \cellcolor[gray]{.8} \ul{46.21} & \cellcolor[gray]{.8} \ul{43.34} & \cellcolor[gray]{.8} 41.02 & \cellcolor[gray]{.8} \textbf{43.91} & \cellcolor[gray]{.8} \ul{33.40} & \cellcolor[gray]{.8} \ul{40.08} & \cellcolor[gray]{.8} \textbf{36.13} & \cellcolor[gray]{.8} \ul{39.30} & \cellcolor[gray]{.8} 30.44 & \cellcolor[gray]{.8} 28.58 & \cellcolor[gray]{.8} \textbf{50.61} & \cellcolor[gray]{.8} \ul{45.89} & \cellcolor[gray]{.8} 43.50 & \cellcolor[gray]{.8} \textbf{27.72} \bigstrut[b]\\
    \hline
    \multicolumn{16}{c}{VGG16} \bigstrut\\
    \hline
    PANet \cite{Wang_Liew_Zou_Zhou_Feng_2019} & 20.05 & 37.71 & 21.18 & 41.22 & 14.15 & 12.17 & 13.82 & 21.05 & 7.89 & 17.88 & 4.36 & 31.68 & 27.55 & 26.88 & 12.97 \bigstrut[t]\\
    CANet \cite{Zhang_Lin_Liu_Yao_Shen_2019} & 24.13 & 6.73 & 13.83 & 16.32 & 8.54  & 14.12 & 3.24 & 21.04 & 3.35 & 22.96 & 9.57 & 14.91 & 17.83 & 16.11 & 9.92 \\
    SCL \cite{Zhang_Xiao_Qin_2021} & 28.50 & 32.93 & 19.68 & 29.60 & 18.05 & 22.48 & 7.92 & 31.46 & 8.99 & 22.02 & 14.17 & 16.53 & 19.72 & \textbf{39.40} & 21.37 \\
    PFENet \cite{Tian_Zhao_Shu_Yang_Li_Jia_2022} & 34.32 & 31.81 & 24.20 & 35.43 & 16.86 & 13.98 & 6.01 & 31.68 & 6.76 & \textbf{26.85} & 8.15 & 17.75 & 20.56 & 33.34 & 14.87 \\
    NERTNet \cite{Liu_Liu_Cao_Yao_Han_Shao} & 12.66 & 23.11 & 26.90 & \ul{50.47} & 15.77 & 23.14 & 8.48 & 31.73 & 11.75 & \ul{24.94} & 14.64 & 20.45 & 29.03 & 28.06 & 7.24 \\
    DCP \cite{Lang_Tu_Cheng_Han_2022} & 27.69 & 38.45 & 25.92 & 33.20 & 15.57 & 17.62 & 12.36 & 26.79 & 8.05 & 17.80 & 22.45 & 18.29 & 18.03 & \ul{37.57} & 16.10 \\
    BAM \cite{Lang_Cheng_Tu_Han} & 27.66 & \ul{43.90} & 31.48 & 43.96 & 22.66 & 13.57 & 8.91 & 31.76 & 9.26 & 20.91 & 17.05 & 26.27 & 30.68 & 25.27 & 8.07 \\
    DMML \cite{Wang_Wang_Sun_Wang_Fu_2022} & 34.75 & 37.36 & 15.15 & 22.85 & 11.94 & 21.41 & 13.85 & 23.92 & 10.24 & 23.50 & 8.17 & 16.32 & 21.08 & 29.63 & 22.09 \\
    DML \cite{Jiang_Zhou_Li_2022} & 27.30 & 42.63 & 19.25 & \textbf{50.63} & 15.13 & 14.16 & 15.94 & 22.40 & 7.74 & 12.74 & 3.79 & 23.73 & 23.47 & 27.40 & 16.88 \\
    SDM \cite{Xie_Liu_Xiong_Shao_2021} & 33.76 & 23.88 & 17.80 & 27.76 & 19.38 & 18.36 & 9.63 & 25.24 & 8.63 & 19.69 & 10.56 & 15.36 & 24.76 & 32.30 & 22.06 \\
    SD-AANet \cite{Zhao_Liu_Lyu_Wang_Yang_2021} & 18.72 & 32.01 & 40.17 & 23.59 & 25.58 & 13.79 & \textbf{48.74} & 13.67 & 14.93 & 15.79 & 8.36 & 22.03 & 18.86 & 11.05 & \ul{23.66} \\
    TBPN \cite{P_Verma} & 22.03 & 39.75 & 20.80 & 42.80 & 13.94 & 10.41 & 6.87  & 16.54 & 4.38 & 23.41 & 5.68 & 23.66 & 22.13 & 24.63 & 14.72 \\
    HMRE \cite{10182361} & 30.25 & 36.16 & \textbf{41.88} & 43.37 & \ul{30.74} & \ul{31.52} & 36.66 & \textbf{35.10} & \ul{18.38} & 19.74 & \ul{29.74} & \textbf{45.09} & \textbf{49.22} & 30.82 & 21.93 \\
    R$^{2}$Net \cite{Lang_Cheng_Tu_Han_2023} & \textbf{37.82} & \textbf{45.16} & 26.27 & 45.30 & 21.81 & 24.11 & 14.38 & 30.92 & 12.21 & 18.03 & 18.66 & 25.02 & 29.64 & 31.95 & 17.87 \\
    \cellcolor[gray]{.8} HSE(\textit{ours}) & \cellcolor[gray]{.8} \ul{37.56} & \cellcolor[gray]{.8} 40.21 & \cellcolor[gray]{.8} \ul{40.65} & \cellcolor[gray]{.8} 36.20 & \cellcolor[gray]{.8} \textbf{31.28} & \cellcolor[gray]{.8} \textbf{34.30} & \cellcolor[gray]{.8} \ul{38.24} & \cellcolor[gray]{.8} \ul{33.26} & \cellcolor[gray]{.8} \textbf{20.01} & \cellcolor[gray]{.8} 23.14 & \cellcolor[gray]{.8} \textbf{32.70} & \cellcolor[gray]{.8} \ul{43.45} & \cellcolor[gray]{.8} \ul{47.63} & \cellcolor[gray]{.8} 30.89 & \cellcolor[gray]{.8} \textbf{25.53} \bigstrut[b]\\
    \hline
    \end{tabular}}%
  \label{tab_3}%
\end{table*}%

\begin{table}[htbp]
  \centering
  \caption{Ablation Study for Each Component of HSE under the $1$-Shot Setting, with the Best Results Highlighted in Bold.}
  \setlength{\tabcolsep}{1mm}{
    \begin{tabular}{c|c|cccc}
    \hline
    \multirow{2}[4]{*}{Backbone} & \multirow{2}[4]{*}{Methods} & \multicolumn{4}{c}{1-shot} \bigstrut\\
\cline{3-6}          &       & Split0 & Split1 & Split2 & Mean \bigstrut\\
    \hline
    \multirow{4}[2]{*}{ResNet50} & Baseline & 39.25 & 31.88 & 35.06 & 35.40 \bigstrut[t]\\          & Baseline+SDI & 43.55 & 35.20 & 37.98 & 38.91 \\
          & Baseline+GCM & 41.01 & 33.17 & 36.66 & 36.95 \\
          & Baseline+GCM+SDI (\textit{ours}) & \textbf{44.03} & \textbf{35.87} & \textbf{39.26} & \textbf{39.72} \bigstrut[b]\\
    \hline
    \end{tabular}}%
  \label{tab_4}%
\end{table}%

\begin{table}[htbp]
  \centering
  \caption{Impact of Different Language Models on the Quality of General Semantic Knowledge under the $1$-Shot Setting.}
  \setlength{\tabcolsep}{2.7mm}{
    \begin{tabular}{c|cccc}
    \hline
    \multicolumn{1}{c|}{\multirow{2}[4]{*}{Language model}} & \multicolumn{4}{c}{1-shot} \bigstrut\\
\cline{2-5}          & Split0 & Split1 & Split2 & Mean \bigstrut\\
    \hline
    SBERT & 43.58 & 35.26 & 38.51 & 39.12 \bigstrut[t]\\
    GloVe & 43.65 & \textbf{35.90} & 39.14 & 39.56 \\
    CLIP  & \textbf{44.03} & 35.87 & \textbf{39.26} & \textbf{39.72} \bigstrut[b]\\
    \hline
    \end{tabular}}%
  \label{tab_5}%
\end{table}%

\begin{table}[htbp]
  \centering
  \caption{Impact of Different Projector Structures on Segmentation Performance. under the $1$-Shot Setting.}
  \setlength{\tabcolsep}{2.7mm}{
    \begin{tabular}{c|cccc}
    \hline
    \multirow{2}[4]{*}{Projector} & \multicolumn{4}{c}{1-shot} \bigstrut\\
\cline{2-5}          & Split0 & Split1 & Split2 & Mean \bigstrut\\
    \hline
    Linear & 44.03 & \textbf{35.87} & 39.26 & 39.72 \bigstrut[t]\\
    2-layers-MLP & \textbf{44.15} & 35.73 & \textbf{39.45} & \textbf{39.78} \\
    3-layers-MLP & 43.34 & 34.26 & 37.52 & 38.37 \bigstrut[b]\\
    \hline
    \end{tabular}}%
  \label{tab_6}%
\end{table}%

\begin{figure}[!]
\centering
\includegraphics[width=3.3in]{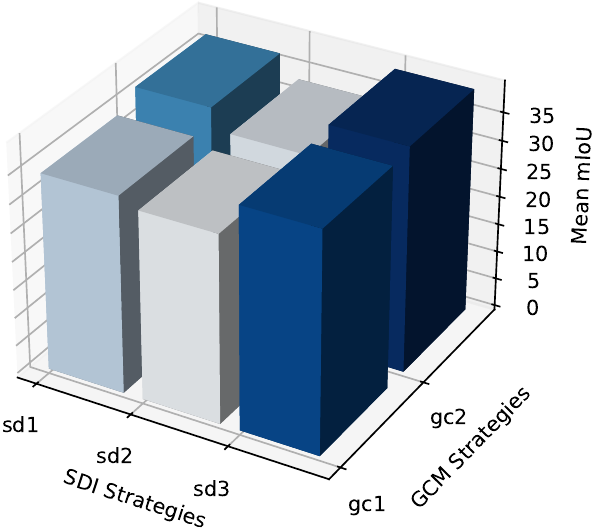}
\caption{Ablation studies on the different designs of SDI and GCM modules under the $1$-shot setting.}
\label{fig_9}
\end{figure}

\begin{figure}[!]
\centering
\includegraphics[width=3.4in]{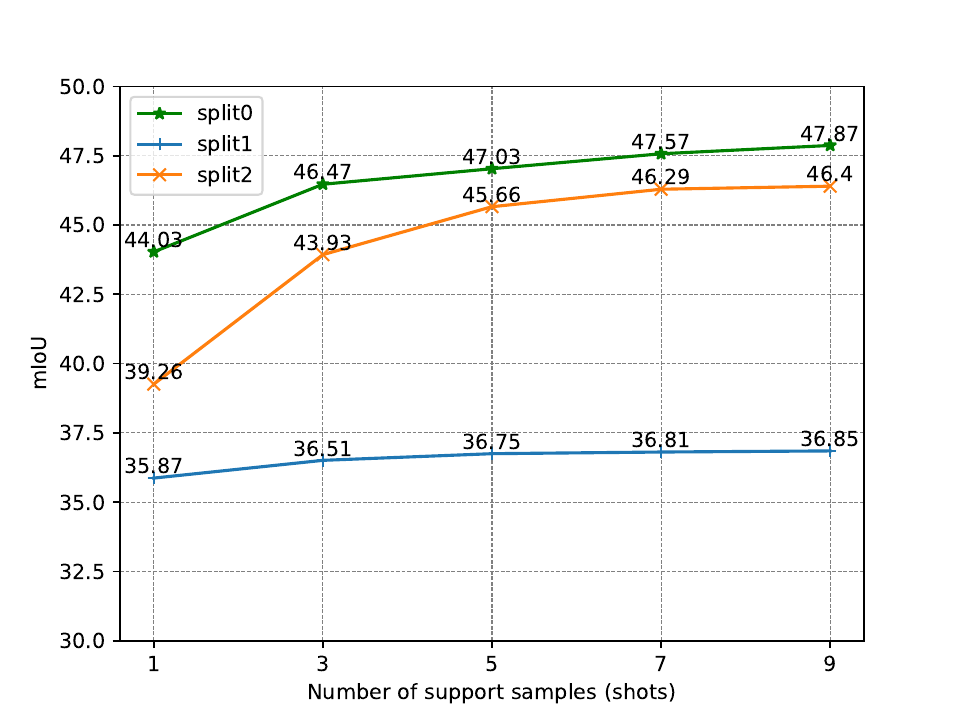}
\caption{Impact of the number of support samples on segmentation performance under the $1$-shot setting.}
\label{fig_9}
\end{figure}

\subsection{Ablation Studies}
\label{Ablation_Studies}

\subsubsection{Component Analysis}
To verify the effectiveness of each module within the proposed HSE framework, we organize component-level ablation experiments.
As shown in Table \ref{tab_4}, the SDI and GCM modules contribute significantly to performance, each enhancing mean mIoU by 3.51\% and 1.55\% respectively. 
This demonstrates that embedding general semantic knowledge from both spatial dense and global content perspectives is beneficial for establishing superior segmentation guidance.
More importantly, the two modules complement each other, synergizing to enhance the performance of the proposed HSE further.

\subsubsection{Language Model Selection}
To explore the impact of general semantic knowledge produced by different language models on segmentation performance, Table \ref{tab_5} displays the results using pre-trained SBERT, GloVe, and CLIP models respectively in the $1$-shot scenario.
It is evident that HSE effectively accommodates different language models and consistently achieves excellent segmentation performance.
Overall, CLIP holds a slight advantage, likely because its pre-training regime better aligns visual and linguistic concepts.
We select CLIP as the default language model in other experiments.

\subsubsection{Projector Structure}
In the design of the SDI and GCM modules, CD embeddings $t$ are mapped to $t^{channel}$ and $t^{spatial}$ using $\mathcal{P}_{channel}(\cdot)$ and $\mathcal{P}_{spatial}(\cdot)$ respectively. 
To identify the optimal projector, comparative experiments are conducted with configurations of \textit{Linear}, \textit{2-layers-MLP}, and \textit{3-layers-MLP}.
From Table \ref{tab_6}, it can be observed that the structure of the projector has a minimal impact on the model's segmentation performance.
Although the \textit{2-layers-MLP} exhibits a slight advantage, considering the overhead of model parameters, we choose the \textit{Linear} as the default projector.

\subsubsection{Design of the SDI and GCM Modules}
We investigate the impact of various implementation strategies for the SDI and GCM modules on segmentation performance.
We evaluate three spatial dense interaction strategies (\textit{sd1}, \textit{sd2}, and \textit{sd3}) and two global content modulation strategies (\textit{gc1} and \textit{gc2}).
Specifically, \textit{sd1} represents the addition of the mid-level support feature $f_{s}$ and the projected CD embedding $t^{spatial}$, \textit{sd2} denotes the element-wise multiplication of $f_{s}$ and $t^{spatial}$, and \textit{sd3} involves concatenating $f_{s}$ and $t^{spatial}$ followed by processing with self-attention, as formulated in Section \ref{Spatial_Dense_Interaction}.
\textit{gc1} represents modulating the general prototype $p_{gen}$ and the mid-level query feature $f_{q}$ using only the enhancement coefficient $w$, while \textit{gc2} employs a residual modulation approach, as described in Section \ref{Global_Semantic_Modulation}.
Fig. X displays the results under different combinations of implementation strategies, where the optimal performance is obtained with the combination of \textit{sd3} and \textit{gc2}.
This demonstrates the superiority of the methods proposed in this paper.

\subsubsection{The Number of Support Samples}
In Fig. X, with the ResNet50 backbone, we illustrate the impact of varying numbers of support samples on the segmentation results for the $1$-shot task.
As the number of support samples increases from 1 to 9, the incorporation of richer visual representations leads to a gradual improvement in performance.
However, when the number of support samples exceeds 5, the incremental improvement in segmentation performance becomes marginal. 
This may be due to the redundancy of target-category-relevant information contained in the numerous support samples.
Additionally, as visual representations dominate the segmentation guidance, the benefits brought by the introduced generalized semantic knowledge are gradually diminished.
 
\section{Conclusion}
In this paper, we proposed the HSE method for the FSS task in remote sensing imagery.
Addressing the particularly severe issue of intra-class differences in remote sensing images, we highlight that relying solely on sparse visual support representations is insufficient to establish robust segmentation guidance.
Based on this observation, as the first attempt to introduce general semantic knowledge from CD embeddings into this field, we designed two complementary and sequential modules: SDI and GCM.
They respectively supplement the missing target scene spatial information in sparse visual representations and modulate the global content related to target categories.
Experiments conducted on the iSAID-5$^{i}$ dataset validate the effectiveness of the designs proposed in this paper, while also demonstrating that the introduced HSE method sets a new benchmark for excellence.

\bibliographystyle{IEEEtran}
\bibliography{IEEEabrv, myrefs}

\end{document}